\documentclass{article}
\usepackage{spconf,amsmath,graphicx}
\usepackage{times}
\usepackage{epsfig}
\usepackage{graphicx}
\usepackage{amsmath}
\usepackage{amssymb}
\usepackage{verbatim}
\usepackage{algpseudocode}
\usepackage{algorithm}
\usepackage{lineno}
\usepackage[pagebackref=true,breaklinks=true,letterpaper=true,colorlinks,bookmarks=false]{hyperref}

\DeclareMathOperator*{\argmin}{arg\,min}

\title{Linear Disentangled Representation Learning for Facial Actions}
\name{Xiang Xiang$^1$ and Trac D. Tran$^2$ 
}
\address{$^1$Dept. of Computer Science \quad \quad $^2$Dept. of Electrical \& Computer Engineering\\
Johns Hopkins University, 3400 N. Charles Street, Baltimore, MD 21218, USA \\
}

\begin{document}
%
\maketitle

\begin{abstract}
Limited annotated data available for the recognition of facial expression and action units embarrasses the training of deep networks, which can learn disentangled invariant features.
However, a linear model with just several parameters normally is not demanding in terms of training data.
In this paper, we propose an elegant linear model to untangle confounding factors in challenging realistic multichannel signals such as 2D face videos.
The simple yet powerful model does not rely on huge training data and is natural for recognizing facial actions without explicitly disentangling the identity.
Base on well-understood intuitive linear models such as Sparse Representation based Classification (SRC), previous attempts require a prepossessing of explicit decoupling which is practically inexact.
Instead, we exploit the low-rank property across frames to subtract the underlying neutral faces
which are modeled jointly with sparse representation on the action components with group sparsity enforced.
On the extended Cohn-Kanade dataset (CK+), our one-shot automatic method on raw face videos performs as competitive as
SRC applied on manually prepared action components and performs even better than SRC in terms of true positive rate.
We apply the model to the even more challenging task of facial action unit recognition, verified on the MPI Face Video Database (MPI-VDB) achieving a decent performance. 
All the programs and data have been made publicly available. 
\end{abstract}
%
%
\vspace{-2mm}
\section{Introduction} \label{sec:intro}
\vspace{-2mm}

\begin{figure}
\begin{center}
   \includegraphics[width=1\linewidth]{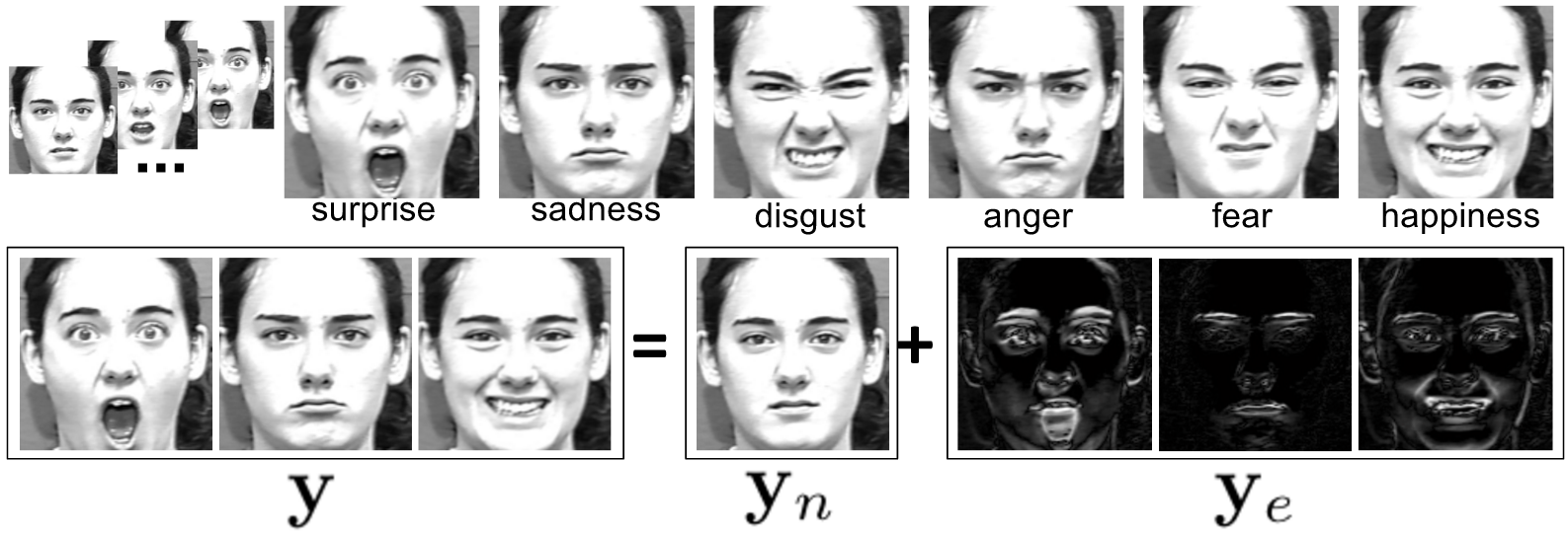}
\end{center}
   \vspace{-8mm}
   \caption{The separability of the neutral face $\mathbf{y}_n$ and expression component $\mathbf{y}_e$. We find $\mathbf{y}_n$ is better for identity recognition than $\mathbf{y}$ and $\mathbf{y}_e$ is better for expression recognition than $\mathbf{y}$.
}
\label{fig:separate}
\end{figure}

\begin{figure}
    \centering
    \includegraphics[width=1\linewidth]{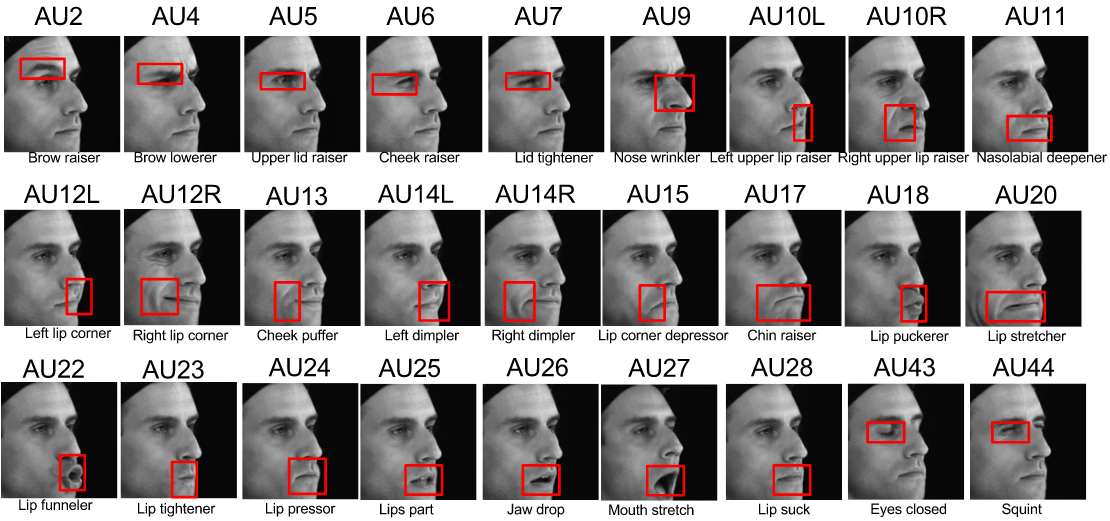}
    \caption{Action unit number and FACS name shown using images from MPI-VDB with 27 distinct AUs. 
    The peak frame is shown. AU12L and AU12R are distinct; similar for AU14.}
    \label{fig:AU_def}
\end{figure}

In this paper, the problem is recognizing facial actions given a face video and action categories in the granularity of either the holistic expression (emotion, see Fig.\ref{fig:separate}) or action units (AU, see Fig. \ref{fig:AU_def}).
The widely-used 6 basic emotions defined by Paul Ekman include \emph{surprise, sadness, disgust, anger, fear} and \emph{happiness}. He also defines the Facial Action Coding System (FACS), using which we can code almost any expression. 
Recently, feature learning \cite{rifai2012disentangling,reed2014learning,liu2014feature} using auto-encoder or adversarial training has shown to be able to disentangle facial \emph{expression} from \emph{identity} and \emph{pose}. 
Unlike face recognition, limited labelled training data are available for facial expressions and AUs in particular.

As shown in Fig.\ref{fig:separate}, an expressive face can be separated into a principal component of the neutral face encoding identity cues and an {\bf action component} encoding motion cues such as the highlighted brow, cheek, lip, lid and nose which relate to AUs in FACS.
As recognition is always broken down into measuring similarity \cite{xiang2016pose}, a similar identity can confuse a similar action. 
To decouple them, \cite{Taheri11} first rules out the neural face explicitly 
and then discriminate between different action components \cite{petrou10} instead of raw faces \cite{ptucha11}. 
The first step is based on the observation that the underlying neutral face stays the same.
If we stack vectors of neutral faces over the time of an action as a matrix, it should be low-rank, ideally with rank $1$. While theoretically the low-rank Principal Component Pursuit \cite{rpca} can be exact under certain conditions, it is of approximate nature in practice.
Their second step is based on the idea of describing an action component as a sparse representation over an over-complete dictionary formed by action components of all categories. 

Our intuition is to remain both facets in a joint manner.
For one thing, we implicitly get rid of the neutral face.
For another, we use equally-spaced sampled frames, since all frames collaboratively yet redundantly represent the expression and neutral face. 
Then, the sparse coefficient vectors form a joint-sparse coefficient matrix.
That drives us to induce both the joint Sparse representation \cite{SRC,eldar10} and the implicit Low-Rank approximation \cite{rpca} in one model (SLR) \cite{xiang2015hierarchical}, which also induces consistent classifications across frames.

Furthermore, ideally non-zero coefficients all drop to the ground-truth category. 
Therefore, the class-level sparsity is $1$ and the coefficient matrix exhibits group sparsity.
However, coefficient vectors share class-wise yet not necessarily atom-wise sparsity \cite{chilasso}.
Thus, we prefer enforcing both the group sparsity and atom-wise sparsity.
We name this extended model Collaborative-Hierarchical Sparse and Low-Rank (C-HiSLR) model \cite{xiang2015hierarchical} following the naming of C-HiLasso \cite{chilasso}.

In \mbox{Sec. \ref{sec:bg}}, we review the classic idea of learning a sparse representation for classification
and its related application on expression recognition. In the remainder,
we first elaborate our model in \mbox{Sec. \ref{sec:model}}, then discuss solving the model via joint optimization in \mbox{Sec. \ref{sec:optm}}, and finally quantitatively evaluate our model in \mbox{Sec. \ref{sec:exp}}, with a conclusion followed in \mbox{Sec. \ref{sec:conclu}}.
\vspace{-4mm}

\section{Related Works} \label{sec:bg}
\vspace{-3mm}
Among non-linear models, one line of work is kernel-based methods \cite{Benitez-Quiroz_2016_CVPR} while another is deep learning \cite{zhao2016peak,Jung_2015_ICCV,liu2014facial,rifai2012disentangling}. 
Similar ideas with disentangling factors have been presented in \cite{liu2014feature,reed2014learning,rifai2012disentangling}.
By introducing extra cues, one line of works is 3D models \cite{Chen_2015_CVPR} while another is multi-modal models \cite{Zhang_2016_CVPR}. But in the linear world, observing a random signal $\mathbf{y}$ for recognition, we just hope to send the classifier a \emph{discriminative compact} representation $\mathbf{x}$ over a dictionary $\mathbf{D}$ such that $\mathbf{Dx=y}$. Normally $\mathbf{x}$ is computed by pursuing the best \emph{reconstruction}.
For example, when $\mathbf{D}$ is under-complete (skinny),
a closed-form approximate solution can be obtained by
Least-Squares: \\
\indent \indent $\mathbf{x}^*=\argmin_{\mathbf{x}} \|\mathbf{y-Dx}\|_2^2 \approx (\mathbf{ D}^T \mathbf{D})^{-1} \mathbf{D}^T \mathbf{y}$. \\
When $\mathbf{D}$ is over-complete (fat),
add a Tikhonov regularizer:
\mbox{$\mathbf{x}^* = \argmin_{\mathbf{x}} \|\mathbf{y-Dx}\|_2^2+\lambda_r \|\mathbf{x}\|_2^2
= \argmin_{\mathbf{x}} \|\mathbf{\widetilde{y}-\widetilde{D}x}\|_2^2$} \\
where $\mathbf{\widetilde{y}=[y,0]}^T$ and $\mathbf{\widetilde{D}=[D},\sqrt{\lambda_r}\mathbf{I]}^T$ 
is under-complete.
Notably, $\mathbf{x}^*=(\mathbf{D}^T \mathbf{D}+\lambda_r \mathbf{I})^{-1} \mathbf{D}^T \mathbf{y}$ is generally dense.

Alternatively, we can seek a sparse usage of $\mathbf{D}$.
Sparse Representation based Classification \cite{SRC} (SRC) expresses a test sample $\mathbf{y}$
as a weighted linear combination $\mathbf{y=Dx}$ of \emph{training samples} simply stacked columnwise in the dictionary $\mathbf{D}$.
Presumably, non-zero weight coefficients drop to the ground-truth class, which induces a sparse coefficient vector or the so-called sparse representation. 
In practice, non-zero coefficients also drop to other classes due to noises and correlations among classes.
Once adding an error term $\mathbf{e}$, we can form an dictionary $\widetilde{\mathbf{D}}$ which is always over-complete: \\
\indent \indent \indent $[\mathbf{x}^*,\mathbf{e}^*]^T = \argmin_{\widetilde{\mathbf{x}}}  sparsity(\widetilde{\mathbf{x}})$\\
\indent \emph{s.t.} \indent $\mathbf{y=Dx+e}=[\hspace{0.2em} \mathbf{D} \hspace{0.3em}|\hspace{0.3em} \mathbf{I} \hspace{0.2em}] \times \left[ \begin{array}{c} \mathbf{x} \\ \mathbf{e}\end{array} \right] = \widetilde{\mathbf{D}}\widetilde{\mathbf{x}}$. \\
SRC evaluates which class leads to the minimum reconstruction error,
which can be seen as a max-margin classifier.

Particularly for facial actions, we treat videos as multichannel signals \cite{eldar10,Zhao_2015_CVPR}, different from image-based methods \cite{Taheri11,petrou10}.
\cite{Taheri11} explicitly separates the neutral face and action component, and then exploits the class-wise sparsity separately for the recognition of identity from neutral faces and  expression from action components.
Differently, with the focus of facial actions we exploit the low-rank property for disentangling identity as well as structured sparsity by inter-channel observation. 
Furthermore, there is tradeoff between simplicity and performance. 
As videos are sequential signals, 
the above appearance-based methods including ours cannot model the dynamics given by a temporal model \cite{Dapogny_2015_ICCV} or spatio-temporal models \cite{liu2014learning,wang2013capturing,guo2012dynamic}. 
Other linear models include ordinal regression \cite{Zhao_2016_CVPR,rudovic2012multi,kim2010structured} and boosting \cite{yang2010exploring}.

   \vspace{-3mm}
\section{Linear Representation Model} \label{sec:model}
   \vspace{-2mm}
In this section, we explain how to model $\mathbf{X}$ using $\mathbf{Y}$ and training data $\mathbf{D}$,
which contains $K \in \mathbb{Z^+}$ types of {\bf actions}.
We would like to classify a test video as one of the $K$ classes.
   \vspace{-8mm}
\subsection{SLR: joint Sparse representation and Low-Rankness} \label{sec:SLR}
\vspace{-2mm}
First of all, we need an explicit representation $\mathbf{Y}$ of an expressive face.
The matrix $\mathbf{Y} \in \mathbb{R}^{d \times \tau}$ can be an arrangement
of $d$-dimensional feature vectors
$\mathbf{y} \in \mathbb{R}^{d}$ ($i=1,2,...,\tau$) of the $\tau$ frames:
$\mathbf{Y} = \big[\mathbf{Y}_1|...|\mathbf{Y}_{\tau} \big]_{d \times \tau}$.
We emphasize our model's power by simply using the raw pixel intensities.

\begin{figure}
\begin{center}
   \includegraphics[width=0.97\linewidth]{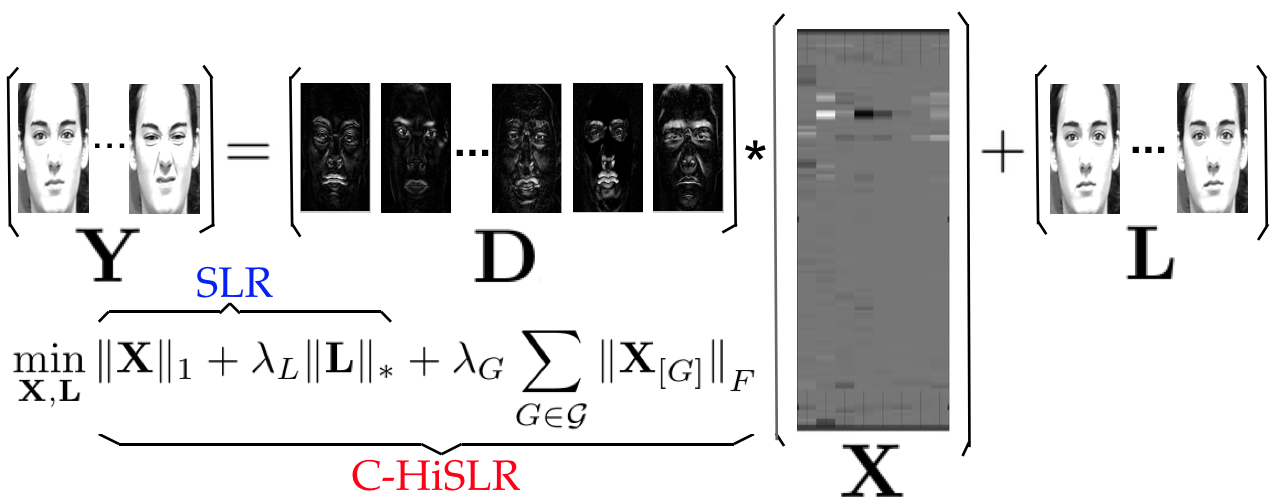}
\end{center}
   \vspace{-8mm}
   \caption{Pictorial illustration of the linear constraint of the proposed model shown for \emph{disgust}. $\mathbf{D}$ is prepared and simply fixed. Depending on the objectve the model has two versions.}
\label{fig:constraint}
\end{figure}

Now, we seek an implicit latent representation $\mathbf{X} \in \mathbb{R}^{n \times \tau}$ of an input test face's emotion $\mathbf{ Y}_e \in \mathbb{R}^{d \times \tau}$
as a sparse linear combination of prepared fixed training emotions $\mathbf{D} \in \mathbb{R}^{d \times n}$:\\
\indent \indent \indent \indent \indent \indent
$\mathbf{Y}_e = \mathbf{DX}$. \\
Since an expressive face $\mathbf{y}=\mathbf{y}_e+\mathbf{y}_n$ is a superposition of an emotion $\mathbf{y}_e \in \mathbb{R}^d$ and a neutral face $\mathbf{y}_n \in \mathbb{R}^d$, we have \\
\indent \indent \indent \indent \indent \indent
$\mathbf{Y} = \mathbf{Y}_e + \mathbf{L}$, \\
where $\mathbf{L} \in \mathbb{R}^{d \times \tau}$ is ideally $\tau$-times repetition of the column vector of a neutral face $\mathbf{y}_n \in \mathbb{R}^d$. Presumably $\mathbf{L} = \big[ \mathbf{y}_n|...|\mathbf{y}_n \big]_{d \times \tau}$. As shown in Fig. \ref{fig:constraint}, $\mathbf{X}$ subjects to \\
\indent \indent \indent \indent \indent \indent
$\mathbf{Y = DX + L}$, \\
where the dictionary matrix $\mathbf{D}_{d \times n}$ is an arrangement of \mbox{all}
sub-matrices $\mathbf{ D}_{[j]}$, $j=1,...,\lfloor \frac{n}{\tau} \rfloor$.
\emph{Only for training}, we have $\lfloor \frac{n}{\tau} \rfloor$ training emotions with neutral faces subtracted.
The above constraint of $\mathbf{X}$ characterizes an affine transformation from the latent representation $\mathbf{X}$ to the observation $\mathbf{Y}$.

If we write $\mathbf{X}$ and $\mathbf{Y}$ in the homogeneous form, we have \\
\indent \indent $\left[ \begin{array}{c} \mathbf{Y}_{d \times \tau} \\ \mathbf{1}_{1 \times \tau} \end{array} \right]
= \begin{bmatrix} \mathbf{D}_{d \times n} & \big(\mathbf{y}_n\big)_{d \times 1} \\ \mathbf{0}_{1 \times n} & 1 \end{bmatrix} \times \left[ \begin{array}{c} \mathbf{X}_{n \times \tau} \\ \mathbf{1}_{1 \times \tau} \end{array} \right]$. \\
In the ideal case with $rank(\mathbf{L})=1$,
if the neutral face $\mathbf{y}_n$ is pre-obtained \cite{petrou10,Taheri11},
it is trival to solve for $\mathbf{X}$.
Normally, $\mathbf{y}_n$ is unknown and $\mathbf{L}$ is not with rank $1$ due to noises.
As $\mathbf{X}$ is supposed to be sparse and $rank(\mathbf{L})$ is expected to be as small as possible (maybe even $1$),
intuitively our objective is to \\
\indent \indent \indent
$\min_{\mathbf{X,L}} sparsity(\mathbf{X}) + \lambda_L \cdot rank(\mathbf{L})$, \\
where
$rank(\mathbf{L})$ can be seen as the sparsity of the vector formed by the singular values of $\mathbf{L}$.
Here $\lambda_L$ is a non-negative weighting parameter we need to tune.
When $\lambda_L = 0$, the optimization problem reduces to that in SRC.
With both terms relaxed to be convex norms, we alternatively solve
\indent \indent \indent \indent \indent
$\min_{\mathbf{X,L}} \| \mathbf{X} \|_1 + \lambda_L \| \mathbf{L} \|_*$,  \\
where $\| \cdot \|_1$ is the entry-wise $\ell_1$ matrix norm,
whereas $\| \cdot \|_*$ is the Schatten $\ell_1$ matrix norm (nuclear norm, trace norm) which can be seen as applying $\ell_1$ norm to the vector of singular values. Now, the proposed joint SLR model is expressed as \\
   \vspace{-2mm}
\begin{equation} \label{eq:slr}
\begin{aligned}
\min_{\mathbf{X,L}} \| \mathbf{X} \|_1 + \lambda_L \| \mathbf{L} \|_*  \indent s.t. \indent \mathbf{Y = DX + L}
\end{aligned}
\end{equation}

\noindent We solve (\ref{eq:slr}) for matrices $\mathbf{X}$ and $\mathbf{L}$ by the Alternating Direction Method of Multipliers (ADMM) (see Sec. \ref{sec:optm}).
   \vspace{-3mm}
\subsection{C-HiSLR: a Collaborative-Hierarchical SLR model} \label{sec:hslr}
   \vspace{-2mm}
If there is no low-rank term $\mathbf{L}$, (\ref{eq:slr}) becomes a problem of multi-channel Lasso (Least Absolute Shrinkage and Selection Operator).
For a single-channel signal, Group Lasso has explored the group structure for Lasso yet does not enforce sparsity within a group,
while Sparse Group Lasso yields an atom-wise sparsity as well as a group sparsity.
Then, \cite{chilasso} extends Sparse Group Lasso to multichannel,
resulting in a Collaborative-Hierarchical Lasso (C-HiLasso) model.
For our problem, we do need $\mathbf{L}$, which induces a Collaborative-Hierarchical Sparse and Low-Rank (C-HiSLR) model:
   \vspace{-2mm}
\begin{equation} \label{eq:hslr}
\begin{aligned}
& \min_{\mathbf{X,L}} \| \mathbf{X} \|_1 + \lambda_L \| \mathbf{L} \|_* + \lambda_G \sum_{G \in \mathcal{G}} {\| \mathbf{X}_{[G]} \|}_F \\
& s.t. \indent \mathbf{Y = DX + L}
\end{aligned}
\end{equation}
   \vspace{-3mm}
\begin{figure}
\begin{center}
   \includegraphics[width=0.72\linewidth]{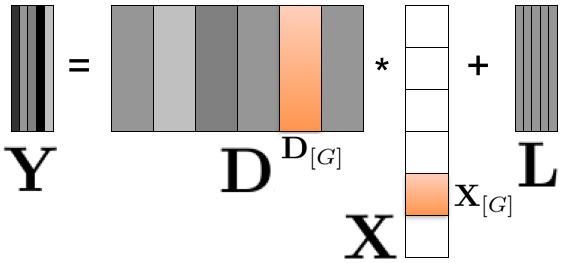}
\end{center}
   \vspace{-8mm}
   \caption{Pictorial illustration of the group sparsity in C-HiSLR.}
\label{fig:gspa}
\end{figure}

\noindent where $\mathbf{X}^{[G]}$ is the sub-matrix formed by all the {\bf rows} indexed by the elements in group $G \subseteq \{1,...,$n$\}$.
As shown in \mbox{Fig. \ref{fig:gspa}},
given a group $G$ of indices, the sub-dictionary of {\bf columns} indexed by $G$ is denoted as $\mathbf{D}_{[G]}$.
$\mathcal{G} = \{G_1,...,G_K\}$ is a non-overlapping partition of $\{1,...,n\}$.
Here ${\| \cdot \|}_F$ denotes the Frobenius norm, which is the entry-wise $\ell_2$ norm as well as the Schatten $\ell_2$ matrix norm and can be seen as a group's magnitude.
$\lambda_G$ is a non-negative weighting parameter for the group regularizer,
which is generalized from an $\ell_1$ regularizer (consider $\mathcal{G}=\big\{\{1\},\{2\},...,\{n\}\big\}$ for singleton groups) \cite{chilasso}.
When $\lambda_G = 0$, C-HiSLR degenerates into SLR.
When $\lambda_L = 0$, we get back to collaborative Sparse Group Lasso.
   \vspace{-5mm}
\subsection{Classification} \label{sec:cls}
   \vspace{-3mm}
Following SRC, for each class $c \in \{1,2,...,K\}$, let $\mathbf{D}_{[G_c]}$ denote the sub-matrix of $\mathbf{D}$ which consists of all the columns of $\mathbf{D}$ that correspond to emotion class $c$
and similarly for $\mathbf{X}^{[G_c]}$.
We classify $\mathbf{Y}$ by assigning it to the class with minimal residual as
  \mbox{$c^* = \argmin_c r_c(\mathbf{Y}) := ||\mathbf{Y} - \mathbf{D}_{[G_c]}\mathbf{X}_{[G_c]} - \mathbf{L}||_F$}.

   \vspace{-4mm}
\section{Optimization} \label{sec:optm}
   \vspace{-3mm}
Both SLR and C-HiSLR models can be seen as solving
   \vspace{-2mm}
\begin{equation} \label{eq:gene-slr}
\begin{aligned}
\min_{\mathbf{X,L}} f(\mathbf{X}) + \lambda_L \| \mathbf{L} \|_* \indent s.t. \indent \mathbf{Y = DX + L}
\end{aligned}
\vspace{-2mm}
\end{equation}

To follow a standard iterative ADMM procedure,
we write down the augmented Lagrangian function for \eqref{eq:gene-slr} as
\vspace{-2mm}
\begin{equation}\label{al}
\begin{split}
&  \mathcal{L}(\mathbf{X},\mathbf{L},\mathbf{\Lambda}) = f(\mathbf{X}) + \lambda_L ||\mathbf{L}||_* \\
&  + \left\langle \mathbf{\Lambda}, \mathbf{Y-DX-L}\right\rangle + \frac{\beta}{2}||\mathbf{Y-DX-L}||_F^2,
\end{split}
\vspace{-5mm}
\end{equation}
where $\mathbf{\Lambda}$ is the matrix of multipliers,
$\left\langle \cdot,\cdot \right\rangle$ is inner product,
and $\beta$ is a positive weighting parameter for the penalty (augmentation).
A single update at the $k$-th iteration includes
{\footnotesize
\begin{align}
  \mathbf{L}_{k+1} &= \argmin_{\mathbf{L}} \lambda_L||\mathbf{L}||_* + \frac{\beta}{2}||\mathbf{Y}-\mathbf{D}\mathbf{X}_k-\mathbf{L}+\frac{1}{\beta}\mathbf{\Lambda}_k||_F^2 \label{LStep} \\
	\mathbf{X}_{k+1} &= \argmin_{\mathbf{X}} f(\mathbf{X}) + \frac{\beta}{2}||\mathbf{Y-DX} -\mathbf{L}_{k+1}+\frac{1}{\beta}\mathbf{\Lambda}_k||_F^2 \label{XStep} \\
	\mathbf{\Lambda}_{k+1} &= \mathbf{\Lambda}_k + \beta(\mathbf{Y}-\mathbf{D}\mathbf{X}_{k+1}-\mathbf{L}_{k+1}). \label{MultStep}
\end{align}}
The sub-step of solving \eqref{LStep} has a closed-form solution:
   \vspace{-3mm}
\begin{equation}\label{LSol}
  \mathbf{L}_{k+1} = \mathcal{D}_\frac{\lambda_L}{\beta}(\mathbf{Y}-\mathbf{DX}_k+\frac{1}{\beta}\mathbf{\Lambda}_k),
  \vspace{-3mm}
\end{equation}
where $\mathcal{D}$ is the shrinkage thresholding operator.
In SLR where $f(\mathbf{X}) = \| \mathbf{X} \|_1$, \eqref{XStep} is a Lasso problem, which we solve by using the Illinois fast solver. When $f(\mathbf{X})$ follows \eqref{eq:hslr} of C-HiSLR, computing $\mathbf{X}_{k+1}$ needs an approximation based on the Taylor expansion at $\mathbf{X}_{k}$ \cite{Minh-rank,chilasso}.
We refer the reader to \cite{chilasso} for the convergence analysis and recovery guarantee.

\vspace{-5mm}
\section{Experimental Results} \label{sec:exp}
\vspace{-2mm}
We evaluate our model on expressions (CK+) and action units (MPI-VDB).
Images are cropped using the Viola-Jones face detector.
Per category accuracies are averaged over 20 runs. 

\vspace{-5mm}
\subsection{Holistic facial expression recognition}
\label{sec:exp-exp}
\begin{figure}
\begin{center}
   \includegraphics[width=0.65\linewidth]{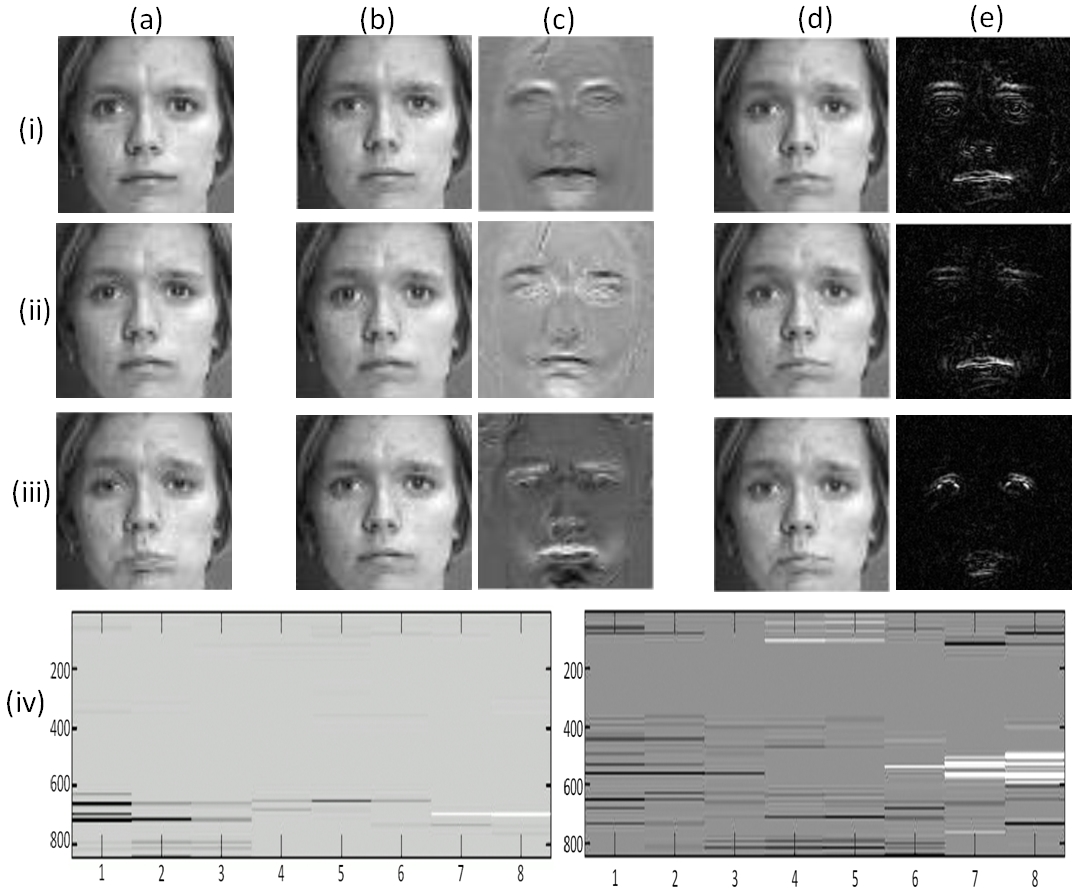}
\end{center}
\vspace{-8mm}
    \caption{Effect of group sparsity. $\tau_{trn}=8$. (a) is the test input $\mathbf{Y}$.
    (b)(c) are recovered $\mathbf{L}$ and $\mathbf{DX}$, given by \mbox{{\bf C-HiSLR}} which correctly classifies (a) as \emph{contempt}.
    (d)(e) are recovery results given by {\bf SLR} which mis-classifies (a) as sadness.
    (i),(ii),(iii) denote results of frame \#1, \#4, \#8 respectively,
    whereas (iv) displays the recovered $\mathbf{X}$
    (left for C-HiSLR and right for SLR).
    $\mathbf{X}$ given by C-HiSLR is group-sparse as we expected.
    }
\label{fig:chi-recovery}
\end{figure}

Experiments are conducted on the CK+ dataset \cite{ck+} consisting of 321 videos with labels\footnote{Contempt is discarded in \cite{Taheri11,petrou10} due to its confusion with anger and disgust but we choose to keep it is for the completeness of the experiment on CK+. See \url{https://github.com/eglxiang/icassp15_emotion} for cropped face data and programs of C-HiSLR, SLR, SRC and Eigenface.}.
For SLR and C-HiSLR, we assume no prior knowledge of the neutral face.
A testing unit contains the last ($\tau_{tst}-1$) frames together with the first frame, 
which is {\bf not} explicitly known a priori as a neutral face.
But for forming the dictionary, we subtract the first frame from the last $\tau_{trn}$ frames per video.
The parameters are set as $\tau_{trn}=8$, $\tau_{tst}=8$, $\lambda_{L}=10$ and $\lambda_{G}=4.5$.
We randomly choose 10 videos for training and 5 for testing per class.
\mbox{Fig. \ref{fig:chi-recovery}} visualizes the recovery results given by C-HiSLR.
Table \ref{tb-gspa} and \ref{tb-jslr} present their confusion matrix, respectively.
Columns are predictions and rows are ground truths. 
Table \ref{tb-sum} summarizes the true positive rate (\emph{i.e.}, sensitivity).
We have anticipated that SLR ({\bf 0.70}) performs worse than SRC ({\bf 0.80}) since SRC is equipped with neutral faces.
However, C-HiSLR's result ({\bf 0.80}) is comparable with SRC's. C-HiSLR performs even better in terms of sensitivity,
which verifies that the group sparsity indeed boosts the performance.

As a comparsion, we replicate the image-based SRC used in \cite{Taheri11,petrou10,ptucha11} and assume the neutral face is provided.
We represent an action component by subtracting the neutral face which is the first frame from the last frame per video.
We choose half of CK+ for training and the other half for testing per class.
When sparsity level is set to 35\%, SRC achieves a recognition rate of {\bf 0.80} shown by Table \ref{tb-src-omp}. 
Accuracies for \emph{fear} \& \emph{sad} are low as they confuse each other.

\vspace{-5mm}
\subsection{Facial action unit recognition}
\label{sec:au-exp}
\begin{figure}
    \centering
    \includegraphics[width=1\linewidth]{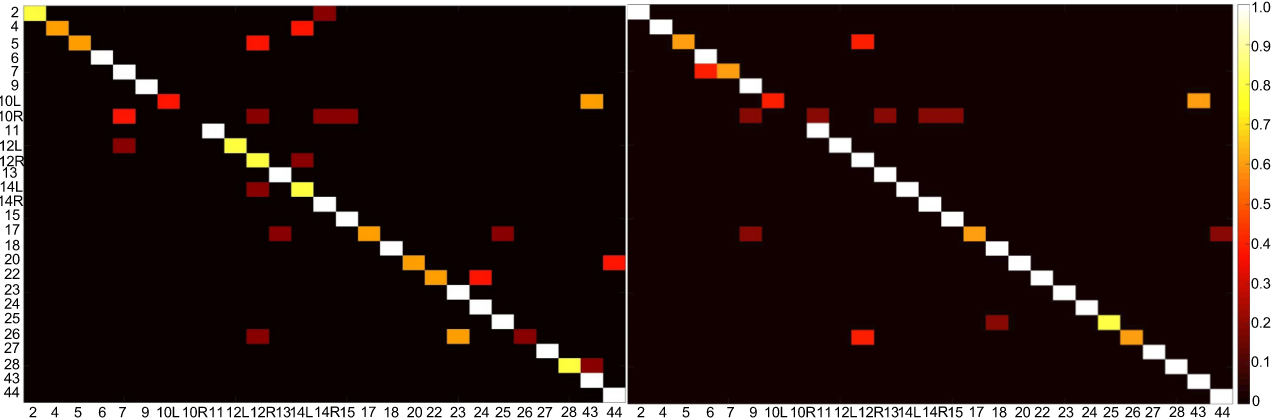}
    \vspace{-4mm}    
    \caption{
    Confusion matrix for {\bf SLR} (left) and {\bf CHi-SLR} (right) on 27 AUs from MPI-VDB. SLR achieves a recognition rate of {\bf 0.80} when the optimizer runs for 30 iters and Lasso run for 3 iters. CHi-SLR achieves a recognition rate of {\bf 0.84} when the optimizer runs for 1000 iters. In both case the std is 0.04.
    }
    \label{fig:AU_SLR}
\end{figure}

To be pose-independant, the following experiments are conducted on a profile view of MPI-VDB \footnote{See \url{http://vdb.kyb.tuebingen.mpg.de} for the raw data and \url{https://github.com/eglxiang/FacialAU} for cropped face data.} containing 27 long video all with over 100 frames (1 video per category, see Fig. \ref{fig:AU_def}). From each video we sample 10 disjoint sub-videos each of which contains 10 equally-spaced sampled frames. 
Different from Sec. \ref{sec:exp-exp}, all frames are directly used without subtracting the first frame as the sub-videos do not start with neutral states.
However, there implicitly exist underlying neural states and presumably the proposed model is still valid.
Then we randomly sample 5 sub-videos from the 10 for training (\emph{i.e.}, forming dictionary) and the other 5 for testing (namely $\tau_{trn}=5$ and $\tau_{tst}=5$).
In this way, the dataset is divided into a training set and a disjoint testing set both with 5 sub-videos per category.
When $\lambda_L = 15$, SLR's performance is shown in Fig. \ref{fig:AU_SLR} with an average recognition rate of {\bf 0.80}.
When $\lambda_G = 4.5$, C-HiSLR's performance is shown in Fig. with a average recognition rate of {\bf 0.84}.
They both perform poorly on AU10R (right upper lip raiser), which confuse with 12R (right lip corner), 13 (cheek puffer), 14R (right dimpler) and 15 (lip corner depressor) because they are all about lips.


\begin{table}
{\small
\begin{center}
\begin{tabular}{|c|c|c|c|c|c|c|c|}
\hline
& An & Co & Di & Fe & Ha & Sa & Su \\
\hline
\hline
An & \bf{0.77} & 0.01 & 0.09 & 0.02 & 0 & 0.07 & 0.04 \\
\hline
Co & 0.08 & \bf{0.84} & 0 & 0 & 0.03 & 0.04 & 0 \\
\hline
Di & 0.05 & 0 & \bf{0.93} & 0.01 & 0.01 & 0.01 & 0 \\
\hline
Fe & 0.09 & 0.01 & 0.03 & \bf{0.53} & 0.12 & 0.07 & 0.15 \\
\hline
Ha & 0.01 & 0.02 & 0.01 & 0.02 & \bf{0.93} & 0 & 0.03 \\
\hline
Sa & 0.19 & 0.02 & 0.02 & 0.05 & 0 & \bf{0.65} & 0.07 \\
\hline
Su & 0 & 0.02 & 0 & 0.02 & 0 & 0.02 & \bf{0.95} \\
\hline
\end{tabular}
\end{center}
\vspace{-6mm}
\caption{Confusion matrix for {\bf C-HiSLR} on CK+ without explicitly knowing neutral faces.
The optimizer runs for 600 iterations and the recognition rate is {\bf 0.80} with a std of 0.05. }
\label{tb-gspa}
}
\end{table}
   \vspace{-2mm}
\begin{table}
{\small
\begin{center}
\begin{tabular}{|c|c|c|c|c|c|c|c|}
\hline
& An & Co & Di & Fe & Ha & Sa & Su \\
\hline
\hline
An & \bf{0.51} & 0 & 0.10 & 0.02 & 0 & 0.31 & 0.06 \\
\hline
Co & 0.03 & \bf{0.63} & 0.03 & 0 & 0.04 & 0.26 & 0.01 \\
\hline
Di & 0.04 & 0 & \bf{0.74} & 0.02 & 0.01 & 0.15 & 0.04 \\
\hline
Fe & 0.08 & 0 & 0.01 & \bf{0.51} & 0.03 & 0.19 & 0.18 \\
\hline
Ha & 0 & 0.01 & 0 & 0.03 & \bf{0.85} & 0.08 & 0.03 \\
\hline
Sa & 0.09 & 0 & 0.04 & 0.04 & 0 & \bf{0.70} & 0.13 \\
\hline
Su & 0 & 0.01 & 0 & 0.02 & 0.01 & 0.02 & \bf{0.94} \\
\hline
\end{tabular}
\end{center}
\vspace{-6mm}
\caption{Confusion matrix for {\bf SLR} on CK+.
We let the optimizer run for 100 iters and Lasso run for 100 iters.
The total recognition rate is {\bf 0.70} with a std of 0.14. }
\label{tb-jslr}
}
\end{table}

\begin{table}
{\small
\begin{center}
\begin{tabular}{|c|c|c|c|c|c|c|c|}
\hline
& An & Co & Di & Fe & Ha & Sa & Su \\
\hline
\hline
An & \bf{0.71} & 0.01 & 0.07 & 0.02 & 0.01 & 0.03 & 0.16 \\
\hline
Co & 0.07 & \bf{0.60} & 0.02 & 0 & 0.16 & 0.03 & 0.12 \\
\hline
Di & 0.04 & 0 & \bf{0.93} & 0.02 & 0.01 & 0 & 0 \\
\hline
Fe & 0.16 & 0 & 0.09 & \bf{0.25} & 0.25 & 0 & 0.26 \\
\hline
Ha & 0.01 & 0 & 0 & 0.01 & \bf{0.96} & 0 & 0.02 \\
\hline
Sa & 0.22 & 0 & 0.13 & 0.01 & 0.04 & \bf{0.24} & 0.35 \\
\hline
Su & 0 & 0.01 & 0 & 0 & 0.01 & 0 & \bf{0.98} \\
\hline
\end{tabular}
\end{center}
\vspace{-6mm}
\caption{Confusion matrix for {\bf SRC} \cite{SRC} on CK+ with neutral faces explicitly provided.
Recognition rate is {\bf 0.80} (std: 0.05).  }
\label{tb-src-omp}
}
\end{table}

\begin{table}
{\small
\begin{center}
\begin{tabular}{|c|c|c|c|c|c|c|c|}
\hline
Model   & An & Co & Di & Fe & Ha & Sa & Su \\
\hline
    SRC & 0.71 & \emph{0.60} & {\bf 0.93} & \emph{0.25} & {\bf 0.96} & \emph{0.24} & {\bf 0.98} \\
\hline
    SLR & \emph{0.51} & \emph{0.63} & \emph{0.74} & {\bf 0.51} & \emph{0.85} & {\bf 0.70} & {\bf \emph{0.94}} \\
\hline
C-HiSLR & {\bf 0.77} & {\bf 0.84} & {\bf 0.93} & {\bf 0.53} & {\bf 0.93} & {\bf 0.65} & {\bf \emph{0.95}} \\
\hline
\end{tabular}
\end{center}
\vspace{-6mm}
\caption{Comparison of sensitivity. The {\bf bold} and \emph{italics} denote the highest and lowest respectively. Difference within $0.05$ is treated as comparable. C-HiSLR performs the best.}
\label{tb-sum}
}
\end{table}
\vspace{-3mm}

\section{Conclusion} \label{sec:conclu}
\vspace{-3mm}
In this paper, we propose a identity-decoupled linear model to learn a facial action representation,
unlike \cite{petrou10} requiring neutral faces as inputs
and \cite{Taheri11} generating labels of the identity and facial action as mutual by-products yet with extra efforts.
Our contribution is two-fold.
First, we do not recover the action component explicitly.
Instead, the video-based sparse representation is jointly modelled with the low-rank property across frames 
so that the neutral face underneath is automatically subtracted.
Second, we preserve the label consistency by enforcing atom-wise as well as group sparsity.
For the CK+ dataset, C-HiSLR's performance on raw faces
is comparable with SRC given neutral faces,
which verifies that action components are automatically separable from raw faces 
as well as sparsely representable by training data.
We also apply the model on recognizing actions units with limited training data, which may embarrass deep learning techniques.

{\footnotesize
\bibliographystyle{IEEEbib}
\bibliography{egbib}
}
\end{document}